\documentclass[letterpaper]{article} 
\usepackage{aaai24}
\usepackage{times}  
\usepackage{helvet}  
\usepackage{courier}  
\usepackage[hyphens]{url}  
\usepackage{graphicx} 
\usepackage{natbib}  
\usepackage{caption} 
\usepackage{algorithm}
\usepackage{algorithmic}
\usepackage{amssymb}
\usepackage{tikz}
\usepackage{amsmath}
\usepackage{diagbox}

\usepackage{newfloat}
\usepackage{listings}
\DeclareCaptionStyle{ruled}{labelfont=normalfont,labelsep=colon,strut=off} 
\floatstyle{ruled}
\newfloat{listing}{tb}{lst}{}
\floatname{listing}{Listing}
\newcommand{\primarykeywords}{
None
}
\usepackage{array}
\usepackage[switch, modulo]{lineno}

\newcommand{\ignore}[1]{{}}
\newcommand{\Ours}{\texttt{BalMCTS}}

\title{$\Ours$: Balancing Objective Function and Search Nodes in MCTS for Constraint Optimization Problems}
\author{
    Yingkai Xiao\textsuperscript{\rm 1},
    Jingjin Liu\textsuperscript{\rm 1},
    Hankz Hankui Zhuo\textsuperscript{\rm 1}$^\ast$
}
\affiliations{
    \textsuperscript{\rm 1}Sun Yat-sen University, Guangzhou, China\\
    xiaoyk8@mail2.sysu.edu.cn, liujj263@mail2.sysu.edu.cn, zhuohank@mail.sysu.edu.cn

}

\usepackage{bibentry}

\begin{document}

\maketitle

\begin{abstract}
Constraint Optimization Problems (COP) pose intricate challenges in combinatorial problems usually addressed through Branch and Bound (B\&B) methods, which involve maintaining priority queues and iteratively selecting branches to search for solutions. However, conventional approaches take a considerable amount of time to find optimal solutions, and it is also crucial to quickly identify a near-optimal feasible solution in a shorter time.
In this paper, we aim to investigate the effectiveness of employing a depth-first search algorithm for solving COP, specifically focusing on identifying optimal or near-optimal solutions within top $n$ solutions. Hence, we propose a novel heuristic neural network algorithm based on MCTS, which, by simultaneously conducting search and training, enables the neural network to effectively serve as a heuristic during Backtracking. Furthermore, our approach incorporates encoding COP problems and utilizing graph neural networks to aggregate information about variables and constraints, offering more appropriate variables for assignments. Experimental results on stochastic COP instances demonstrate that our method identifies feasible solutions with a gap of less than 17.63\% within the initial 5 feasible solutions. Moreover, when applied to attendant Constraint Satisfaction Problem (CSP) instances, our method exhibits a remarkable reduction of less than 5\% in searching nodes compared to state-of-the-art approaches. 
\end{abstract}

\section{Introduction}
Constrained Optimization Problems (COP) \cite{modi2005adopt} represent a universal mathematical paradigm extensively applied in modeling a diverse range of real-world challenges spanning transportation, supply chain, energy, finance, and scheduling \cite{bertsekas2014constrained, rollon2014decomposing}.

Traditionally, the pursuit of global optimality in solving COPs has been anchored in the Branch-and-Bound (B$\&$B) algorithm, which employs a divide-and-conquer strategy. A pivotal aspect of B$\&$B is the branching rule, recursively partitioning COP search space within the search tree. While optimal branching decisions in a general setting remain a subject of limited understanding \cite{lodi2017learning}, the selection of branching rules significantly influences practical solving performance \cite{achterberg2013mixed}. Leading modern solvers achieve state-of-the-art performance by employing hard-coded branching rules grounded in heuristic methods devised by domain experts, showcasing effectiveness across a representative set of test instances \cite{gleixner2021miplib}. Recent studies have been turned to Machine Learning (ML) to obtain effective branching rules and enhance expert-crafted heuristics. Given the sequential nature of branching, Reinforcement Learning (RL) emerges as a natural paradigm for learning, enabling direct searches for optimal branching rules with respect to metrics such as average solving time or final B$\&$B tree size \cite{scavuzzo2022learning, DBLP:journals/eaai/SongC0X022}. However, RL-based approaches often necessitate numerous training iterations for model convergence. 

Apart from this, Belief Propagation (BP) \cite{kschischang2001factor} stands out as a crucial message-passing algorithm for COPs, aiming to quickly find a good feasible solution. Additionally, the Min-sum \cite{farinelli2008decentralised} algorithm is designed to propagate cost information throughout factor graphs, ultimately seeking cost-optimal solutions. It is noteworthy that the conventional BP encounters challenges in ensuring convergence, particularly when dealing with graphs containing cycles.  Nevertheless, the factor graph can attain convergence by employing Damped BP, achieved by blending the newly composed message from each iteration with the old message from the preceding iteration.
Self-supervised DNN-based BP approach is proposed in the state-of-the-art Deep Attentive Belied Propagation (DABP) \cite{deng2022deep}, which is based on dynamic weights and damping factors to obtain optimal solutions for solving COPs. While promising, this approach is still limited by the high costs and convergence times associated with complex problems.
\emph{For instance, as illustrated in Table \ref{tab:task_assignment}, the task is to assign jobs to employees, where each employee incurs different costs associated with various tasks. Our objective is to minimize the total cost by determining the optimal task assignment for each employee. Employing the B$\&$B method to seek the optimal solution entails exploring all possible task assignments for each employee in the first search to obtain the minimum lower bound of Eric, followed by systematically attempting task assignments for the second employee, Mary, and so forth. This exhaustive search approach requires exploring 13 iterations to attain the optimal solution.} We can envision that with the increase in problem size, traditional optimization algorithms may struggle to handle COP effectively, as the expansion of the search space and the complexity of constraints could lead to a decline in algorithm performance. Therefore, in this paper, we aim to address the following two issues: \textbf{1) How to design and implement optimization algorithms suitable for complex constraints? 2) Confronted with intricate problem structures, can we efficiently find the approximate optimal solution within an acceptable time?}

In response to the above-mentioned issues, we propose a novel approach, namely \textbf{Bal}ancing Objective Function (c.f. Eq. (\ref{target})) and Search Nodes in Monte Carlo Tree Search ($\Ours$), which involves training a balancing neural network on MCTS-generated exploration and utilizing the network to guide subsequent MCTS exploration, acting as an effective heuristic for efficient branching rule during the solution process.
\emph{Table \ref{tab:task_assignment} also demonstrates the advantages of our algorithm, utilizing the value assignment strategy of choosing the minimum value. For the initial task assignment, we first select the employee Alex for task 3, Mary for task 2, and Eric for task 4, requiring only three assignments to reach the optimal solution.} This innovative integration of MCTS and the balancing neural network training process offers significant enhancements to COPs, as outlined below:

\begin{itemize}
\item We develop a Monte Carlo Tree Search (MCTS) based constraint optimization approach that strikes a balance between the objective function and the number of search nodes to address COPs, effectively guiding the neural network in learning branching rules through a variant of Upper Confidence Bound for Tree (UCT),  thereby controlling the search direction. Within an acceptable time, we identify more and higher-quality feasible solutions, thus enhancing the performance of the model.
\item Additionally, we enhanced the encoding scheme for COPs by transforming COP instance state into corresponding encodings using Graph Neural Network (GNN). The encoding improved the inference speed of the network, enabling rapid identification of near-optimal solutions in complex COPs.

\item  Finally, we performed multi-objective optimization, aiming to minimize the search nodes and the objective function, which balances multiple conflicting objectives of COPs, allowing for the rapid identification of nearly optimal feasible solutions within a short time.
\end{itemize}

\begin{table}[!ht]
\centering
\scriptsize
\resizebox{\linewidth}{!}{
\begin{tabular}{|c|c|c|c|c|}
     \hline
     & Task 1 & Task 2 & Task 3 & Task 4 \\
     \hline
Eric & 9                      & 5                       & 7                       & \textcolor{red}{6}      \\
     \hline
Mary & 6                      & \textcolor{red}{2}      & 0                       & 7      \\
     \hline
Emma & 5                      & 8                       & \textcolor{red}{1}      & 8      \\
     \hline
Alex & \textcolor{red}{4}     & 6                       & 9                       & 4      \\
     \hline
\end{tabular}
}
\caption{Task Assignment}
\label{tab:task_assignment}
\end{table}

\section{Related Work}
\textbf{MCTS Applications in Combinatorial Optimization.} 
The emergence of AlphaGo \cite{silver2017mastering} as the winner in the Masters, illustrates the success of MCTS in two-player games. Combinatorial optimization problems are sometimes transformed into game-like structures using MCTS. In specific instances, UCT has been used to guide MIP solvers \cite{sabharwal2012guiding} and Quantified Constraint Satisfaction Problems (QCSP) \cite{satomi2011real}. Additionally, Bandit Search for Constraint Programming (BaSCoP) \cite{loth2013bandit} represents another significant instantiation of MCTS, enhancing the tree-search heuristics of Constraint Programming (CP) solvers and exhibiting substantial improvements in depth-first search on certain CP benchmarks. More recent works involve the successful application of the Neural MCTS of AlphaZero to solve a variety of NPHard graph problems \cite{abe2019solving}, as well as solving First-order logic descriptions of combinatorial problems \cite{xu2021first}.

In contrast to the aforementioned approaches, our algorithm diverges in its objective, aiming not for an optimum or a local optimum but rather to explore for a high-quality solution while utilizing as few nodes as possible. This presents a formidable challenge, as seeking a feasible solution with minimal node exploration is inherently a difficult problem, further compounded by the additional complexity of pursuing a high-quality solution.


\section{Preliminaries}
In this section, we briefly present the background for several highly relevant topics.
\subsection{Constraint Optimization Problems}
A Constraint Optimization Problem is formally defined by a triple $(X, D, C)$ where,
\begin{itemize}
\item $X = \{ x_1, x_2, \dots, x_n \}$ denotes the set of variables, where each $x_i$ corresponds to an element to be solved in the problem domain.
\item $D = \{ d_1, d_2, \dots, d_n \}$ denotes the set of domains, where each domain set $d_i$ contains the possible values that variable $x_i$ can take.
\item $C = \{ c_1, c_2, \dots, c_m \}$ denotes the set of constraint functions, each constraint function $c_j$, where the scope $\langle scp(c_j)$, $rel(c_j) \rangle$, define the cost associated with each potential assignment of variables within their scope.
$scp(c_j) \subseteq X$ denotes the scope of constraint function $c_j$ by specifying which involved variables, while $rel(c_j)$ represents a relation containing permissible value combinations of the variables in $scp(c_j)$ and corresponds to their associated weights.
\end{itemize}

\begin{figure}[t]
\begin{minipage}[t]{0.5\textwidth}
\centering
\includegraphics[width=1.0\linewidth]{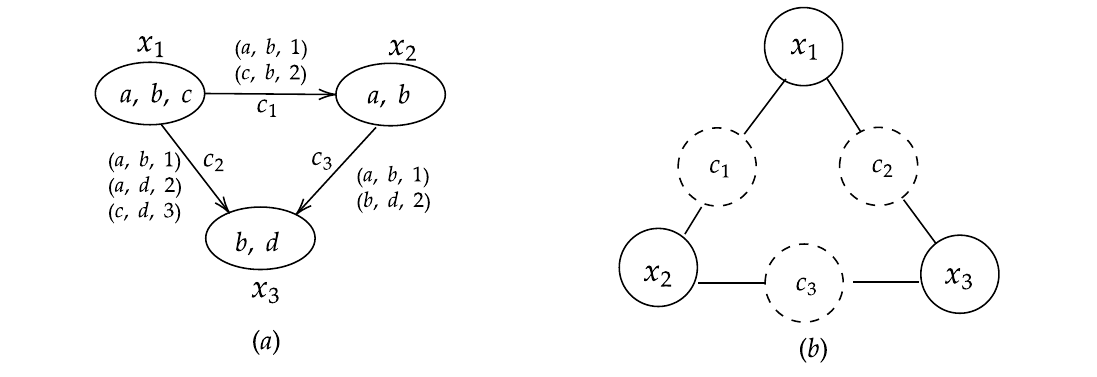} 
\caption{An example of a graph for COP instance (state)}
\label{fig:figure1}
\end{minipage}
\end{figure}

\textit{A COP instance can be described as a constraint graph, as shown in Figure \ref{fig:figure1}, where nodes represent variables $X$, and edges represent constraint functions $C$. Specifically, in Figure \ref{fig:figure1}(a), the domain of variable node $x_1$ is $d_1 = \{a, b, c\}$, and that of $x_2$ is $d_2 = \{a, b\}$. The constraint function $c_1$ represents an edge from $x_1$ to $x_2$, where the scope of $c_1$ is $scp(c_1) = \{x_1, x_2\}$ and relation is $rel(c_1) = \{(a, b, 2), (c, b, 2)\}$. Furthermore, $rel(c_1)$ indicates possible value combinations for variables $x_1$ and $x_2$ as $(a, b)$ and $(c, b)$, with associated correlation weights of 2. As $x_1$ and $x_2$ belong to $scp(c_1)$, $c_1$ transforms into a constraint node, establishing corresponding edges with variables $x_1$ and $x_2$ in Figure \ref{fig:figure1}(b).}

\textbf{Objective Function} The goal of COP is to determine an assignment of values to variables, such that all the given constraints $c_j \in C$ are satisfied and minimize the following objective function $\tau$:
\begin{equation}\label{target}
    \tau = min \sum_{c_j \in C} Weight(\eta | rel(c_j))
\end{equation}
where $\eta$ belongs to the $rel(c_j)$ and $Weight(\eta$) represent the weight of $\eta$.

\subsection{State and Action Space}

\emph{\textbf{Definition 1 (State)} We define the state as a graph structure representing a Constraint Optimization Problem (COP) instance, denoted as the COP instance graph. This graph consists of variables and constraints nodes. Notely, the neighbours of each constraint node are variable nodes, and the neighbors of each variable node are constraint nodes. Specifically, the state is an instance for COP defined as $s = \{x_1,x_2,...,x_n; d_1, d_2, ..., d_n; c_1,...,c_j\}$, where $x_i$ represents the $i$-th variable node, $d_i$ represents the feasible domain for $x_i$, and $c_j$ represents the $j$-th constraint node. } \\

\emph{\textbf{Definition 2 (Action)} We define $a$ as operations that select a variable. By assigning a value to the selected variable, the information of nodes in the COP instance graph is modified, leading to a transition from the current state $s$ to a new state $s'$. This state transition is based on operations such as constraint propagation performed on the selected variable. The set of actions is defined as $A(s)=\{a_1, a_2, ..., a_k\} \subseteq \{a_1, a_2, ..., a_n\}$, where $k$ is the number of variables that are not yet assigned values in the state $s$ and $a_i$ represents the action of selecting the $i$-th unassigned variable.}

\section{Method}

\begin{figure*}[!ht]
\centering
\includegraphics[width=1.1\linewidth]{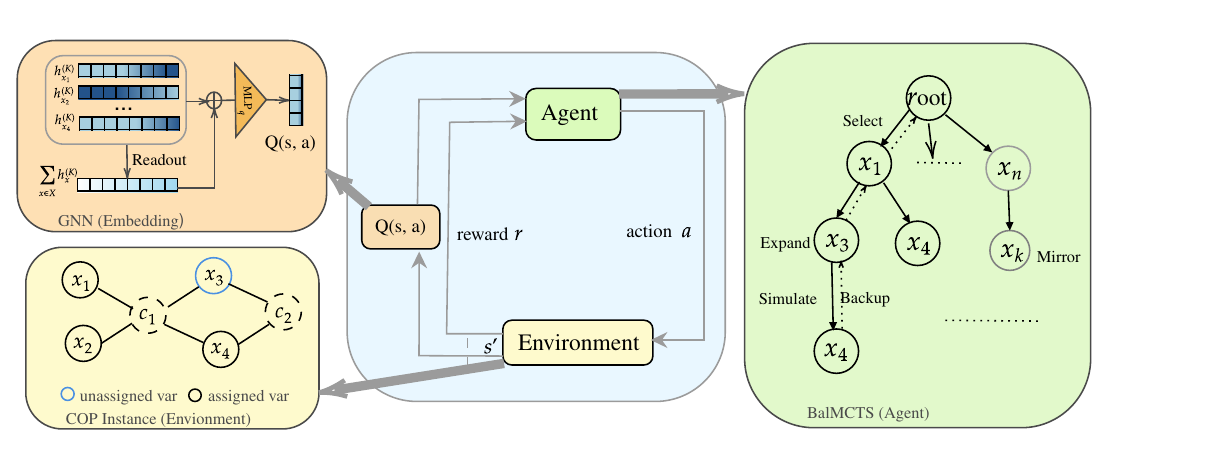} 
\caption{Overview of our $\Ours$ approach}
\label{fig:figure1a}
\end{figure*}



In this section, we provide a detailed explanation of the steps involved in training a network based on MCTS. 
Figure \ref{fig:figure1a} illustrates the overarching framework of our algorithm, elucidating the collaborative interplay of its constituent modules during the training phase. The COP Instance (Environment) module, utilized for depicting the current state of the COP instance, employs graphical representation and undergoes encoding of graph nodes through the Graph Neural Network (GNN) module. Consequently, this process yields the $Q(s, a | a \in A(s))$ for the current state $s$. Within the $\Ours$ (agent), $Q(s, a)$ serves as an integral component of the UCT variant formula, facilitating the selection of the action $a$ associated with the minimal score within the state $s$. After the execution of action $a$, it reaches a new state node $s'$. In this new state $s'$, a fresh round of action selection ensues, representing the new state $s'$ as a COP instance graph. Iterating through this process enables MCTS to discover a plethora of high-quality feasible solutions, thereby effectively training the GNN module. This, in turn, enhances the branching capabilities of the network during the problem-solving phase.
Subsequently, we present our training algorithm.
\subsection{GNN-based Representation}

To adopt the neural network for variable recommendation in the backtracking search, we first need to encode the search states. As mentioned above COP, a COP instance can be represented as a constraint graph as shown in Figure\ref{fig:figure1}(a), where the constraint graph serves as a state $s$, with nodes representing variables and edges representing constraint functions.
To enhance the representation, constraint functions can also be expressed as nodes, as shown in Figure \ref{fig:figure1}(b). In this way, it becomes possible to embed nodes of the constraint graph using GNN. In this paper, we adopt the embedding method proposed by \cite{DBLP:conf/iclr/XuHLJ19} and \cite{DBLP:journals/eaai/SongC0X022}, leveraging the power of GNN as a powerful framework for learning expressive vector representations to aggregate information from neighboring nodes. The description of the GNN variant is provided as follows.

In this GNN variant, we use the notation $h_x^{(k)}$ to represent the $p$-dimensional embedding of variable $x$ in the $k$-th layer, and $h_c^{(k)}$ to represent the $p$-dimensional embedding of constraint function $c$ in the $k$-th layer. $\mathcal{N}(x)$ and $\mathcal{N}(c)$ denote the sets of neighboring nodes for variable $x$ and constraint function $c$, respectively. Additionally, we use $F_x$ and $F_c$ to represent the raw features of variable node $x$ and constraint node $c$, where vector dimensions are $p_x$ and $p_c$ correspondingly. The raw features $F_x$ for variable nodes include the current domain size and variable binding status, thus $p_x = 2$. For constraint nodes, the raw features $F_c$ comprise three dimensions: 1) the number of bound variables in the constraint, 2) the dynamic tightness \cite{li2016improving} of the constraint (a heuristic approach), and 3) the ratio between the current minimum attainable value and the maximum value over the entire domain, therefore, $p_c = 3$.
More specifically, the detailed steps of iterative aggregation and updating of node features for the GNN variant are shown below:
\begin{equation}
    h_x^{(0)} = F_xw_x, \ \  h_c^{(0)} = F_cw_c \label{eq:init}
\end{equation} 

\begin{equation}
    h_x^{(k)} = MLP_x \left(\sum_{c \in \mathcal{N}(x)}h_c^{(k)} : h_x^{(k - 1)} : F_x \right) \label{eq:aggregateVariable}
\end{equation}

\begin{equation}
    h_c^{(k)} = MLP_c \left(\sum_{x \in \mathcal{N}(c)}h_x^{(k - 1)} : h_c^{(k - 1)} : F_c \right) \label{eq:aggregateConstraint}
\end{equation}

Here, the notation $(-:-)$ denotes the concatenation operator. Therefore, the input and output dimensions of $MLP_x$ are $2p+p_x$ and $p$ respectively, while those of $MLP_c$ are $2p+p_c$ and $p$. The Eq.(\ref{eq:aggregateVariable}) and Eq.(\ref{eq:aggregateConstraint}) employ MLP to iteratively aggregate information and update node features for variable node $x$ and constraint node $c$ at $k-1$ layer, respectively. Eq.(\ref{eq:init}) is utilized for parameter initialization, where $w_x \in \mathbb{R}^{p_x \times p}$ and $w_c \in \mathbb{R}^{p_c \times p}$ are learnable parameters. 

Based on the GNN variant, we obtained the representation $h_x^{(K)}$ for each variable by $K$ iterative aggregation, which allows for future embedding of COP instances. To represent the state $s$, which corresponds to different graph representations, we introduced the action-value function $Q(s,a)$. Here $Q(s, a)$ indicates the Q-value after taking action $a$ in the current state $s$ (i.e., selecting a variable). For example, in the current state $s$ where three variables $\{x_1,x_2, x_3\}$ are yet to be assigned values, corresponding to three available actions, we choose the action $a = \text{argmin}_{a \in A(s)} Q(s, a)$ that minimizes the Q-value. Specifically, we denote the current state $s$ by conducting graph-level pooling, achieved through element-wise summation of all variable embeddings after $K$ iterations. Subsequently, we concatenate the embedding representations of the graph with the corresponding action $a$, feeding them into another MLP$_q$ to obtain $Q(s,a)$, expressed as follows:

\begin{equation} 
    Q_w(s, a) = MLP_q\left( \sum_{x \in X} h_x^{(K)} : h_a^{(K)} \right) \label{eq:Q(s,a)}
\end{equation}
where the input and output dimensions of MLP$_q$ are $2p$ and $1$ respectively. 
  
\subsection{Monte Carlo Tree Search formulation}


 
\begin{figure}[!ht]
    \begin{minipage}{0.5\textwidth}
    \centering
    \includegraphics[width=1.0\linewidth]{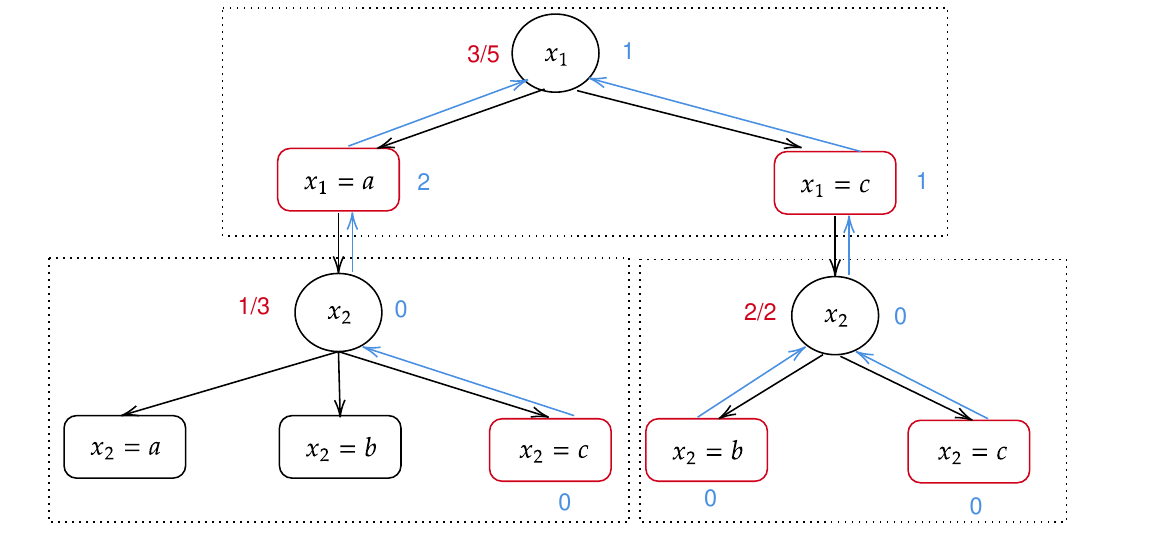} 
    \caption{Backup}
    \label{fig:figure3}
    \end{minipage}
\end{figure}

Monte Carlo Tree Search (MCTS) is a heuristic search algorithm employed in decision-making processes. In addressing Constrained Optimization Problems (COP), we have devised the $\Ours$ method. Within our framework, two types of nodes exist, namely variable nodes and value nodes. Given that each variable encompasses multiple desirable values, corresponding value nodes are associated with each variable. As depicted in Figure \ref{fig:figure3}, the circle nodes represent variable nodes, while square nodes denote specific value nodes for variables. The training process of MCTS consists of five key steps: 1) Selection: we choose a $Q(s, a)$ with the minimum score based on a variant of the UCT formulation until reaching a leaf node (variable node); 2) Expansion: due to the presence of the multiple desirable values associated with leaf nodes, one value node is randomly selected for expansion, generating all child nodes (variable nodes) associated with the chosen value node; 
3) Simulation: conducting a simulation on the expanded child nodes; 4) Mirror: if the simulation results in a feasible solution, a mirroring operation is performed, akin to a data augmentation step; 5) Backup: upon simulation reaching a terminal state, information regarding the terminal state is transmitted back to all nodes along the path, updating the information about the nodes, including success/failure $r(s)$ and objective function $\tau(s)$. This series of steps constitutes the comprehensive process of MCTS, facilitating the discovery of a greater quantity of higher-quality feasible solutions within a reasonable time, thus enhancing the model performance.

In the following sections, we provide a detailed description of how each of these four phases is executed.

\begin{enumerate}

    \item SELECTION: When selecting a node in the MCTS, the selection is based on a variant of the UCT formula that combines the Q-value of the node, denoted as $Q(s, a)$, with an exploration term $U(s, a)$ and a hyperparameter $\alpha(s)$. The formula is given as:
    \begin{equation}
        \pi(s) = \arg\min_{a \in  A(s)}(Q(s, a) - c_1 \times U(s, a) - \alpha(s)) \label{eq:pi}
    \end{equation}
    where $\arg\min_{a \in A(s)}(Q(s, a))$ represents select a child of minimum Q-value. The $U(s, a)$ is defined by the UCT exploration formula:
    \begin{equation}
        U(s, a) = \sqrt \frac{{log(\sum_{a}N(s, a))}}{N(s', a')} 
    \end{equation}
    where $N(s, a)$ represents the number of times action $a$ was taken from state $s$ during exploration. $s'$ is a child node of $s$, and $c_1$ parameter denotes the degree of exploration. As the number of iterations increases, the value of $U(s, a)$ decreases, resulting in more accurate results. The hyperparameter $\alpha(s)$ associated with each node plays a crucial role in controlling the search direction, enabling the algorithm to explore the search space effectively and discover additional solutions. $\alpha(s)$ initialized to 0 and adjusted to $\alpha(s)$ are made during the MIRROR phase.

    \item EXPANSION: When a variable node $x_i$ is selected by SELECTION, and $x_i$ has a corresponding domain $d_i$, a random assignment is made to $x_i$ to obtain a value node $(x_i = j); j \in d_i$. Subsequently, all child nodes of variable node $x_i$ are expanded based on the value nodes $(x_i = j)$. The expansion of nodes is conditioned on the set of variables that are yet to be assigned values.

    \item SIMULATION: The simulation process involves the random selection of unbound variables for value assignment, and repeating until all variables are assigned values or until constraint propagation reveals the impossibility of finding a solution. The termination states are defined as either identifying a solution or conclusively establishing the absence of a solution. Also, the number of simulation assignments is at most equal to the greater depth of the tree,i.e. the number of variables.

    \item MIRROR: It is a pivotal phase of customization in our algorithm that exerts a critical influence on the overall performance and involves a mirroring or copying operation. Primarily, it is imperative to acknowledge a fact, if a solution is represented as $(x_1 = a, x_2 = b, x_3 = c)$, then rearranging the order of variables, as in $(x_2 = b, x_1 = a, x_3 = c)$, still constitutes a valid solution, However, initiating the search from the node $x_2 = b$ may lead to a disparity in the number of search nodes for finding the solution $(x_2 = b, x_1 = a, x_3 = c)$ compared to commencing the search from $x_1 = a$. This discrepancy arises from the distinct variable domains resulting from the constraint propagation initiated by selecting $x_2 = b$ as opposed to $x_1 = a$. Consequently, the probability of finding a solution or not finding a solution when performing the simulation will also be different. 
    Hence, when encountering a solution during simulation, we execute a MIRROR operation. Specifically, we randomly permute the order of the last two variables in the solution, treating it as a new solution, and subsequently expand the nodes corresponding to this new solution. For instance, if there exists a path in the current search tree $(x_1 = a, x_2 = b, x_3 = c)$, we interchange the order of the last two variables, yielding a new path $(x_1 = a, x_3 = c, x_2 = b)$. If there is no sub-node for $x_3$ after $x_1 = a$ along this path, we expand this new node. If this node already exists, we enhance the value of $\alpha(x_1 = a, x_3 = c, x_2)$ to diminish the value of $\pi(x_1 = a, x_3 = c, x_2)$, which steering the MCTS towards this direction. In essence, performing a MIRROR operation is equivalent to performing a data augmentation, thus improving the performance of our model. 
    \item BACKUP: The information of each node is characterized by two attributes, denoted as ($\tau(s), r(s)$), where $\tau(s)$ represents the objective function of state $s$, and $r(s)$ indicates the success rate, computed as the number of successful simulations of state $s$ / the total number of simulation, serving as a metric for the search node count associated with state $s$. Additionally, the objective function $\tau(s)$ reflects the historically best objective function, which means that if the objective function of the current node is lower than the recorded historical value, no updates are made to the $\tau(s)$ of the nodes among the path, otherwise, updates are performed. In the BACKUP phase, two values are propagated. The first value, either success = 1 or failure = 0, signifies the simulation outcome — indicating the generation of a feasible solution or the absence solution. The second value is a solution, enabling the update of $\tau(s)$ for each node along the path from the leaf node to the root node. As depicted in Figure \ref{fig:figure3}, the variable node $x_1$ undergoes 5 simulations with a success rate of 3 out of 5. Therefore, $r(s) = 3/5$, and a similar calculation is applied to other variable nodes. The objective values for $x_1 = a$ and $x_1 = c$ are 2 and 1, respectively. The variable node selects the minimum value among its children, i.e., $\tau(x_1) = \min(\tau(x_1 = j | j \in d_1))$. This approach is mirrored for the $x_2$ node. It is noteworthy that the objective function of the leaf node $x_2 = c$ is 0, given the absence of a cost function for the leaf node. However, if a solution is simulated with a superior objective compared to the existing one, the objective is updated accordingly. It is crucial to emphasize that each subtree maintains its unique objective value.
\end{enumerate}
 
\subsection{Training algorithm}
Our training algorithm $\Ours$ is based on MCTS and Double Deep Q-Network (DDQN). DDQN maintains two networks, namely \emph{online network} $Q_w$ and \emph{target network} $Q_{\bar{w}}$. Specifically, $Q_{\bar{w}}$ is a periodical copy of $Q_w$. In each state of the Selection phase, it utilizes the prediction value $Q_{\bar{w}}(s, a)$ for selecting actions among child nodes of state $s$, and adds tuple $(s, a, Q_{\bar{w}}(s, a))$ to experience buffer $\mathcal{R}$ with size $\mathcal{M}$. Then a mini-batch of $\mathcal{B}$ transitions is sample from $\mathcal{R}$ to updated the parameters of the online network $Q_w$ by performing gradient descent to minimize the squared loss between $Q_w(s, a)$ and the following target:
\begin{equation}
    y = c_3 \times  g(\tau(s')) + c_4 \times (1 - r(s'))
    \label{eq:y}
\end{equation}
where $s'$ represents the state obtained by choosing action $a$ in the state s, denoted as $s' \leftarrow (s, a)$, and $g(\tau(s'))$ denotes the gap of $\tau(s')$. 

\begin{algorithm}[ht]
\caption{Balancing Objective Function and Search Nodes in Monte Carlo Tree Search ($\Ours$)}
\label{alg:algorithm}
\textbf{Input}: COP instances $\mathcal{P}$ \label{alg:input}\\ 
\textbf{Parameter}: $\mathcal{T_{\text{max}}}, \mathcal{N}$ \label{alg:param}\\ 
\textbf{Output}: $\bar{w}$ \label{alg:output}
\begin{algorithmic}[1] 
    \STATE Initialize the experience replay to capacity $\mathcal{M}$ \label{alg:line:initM} \\
    \STATE create root node $x_0$ with state \label{alg:line:root} \\
    \FOR{instance $p \in \mathcal{P}$ } \label{alg:line:p}
    \STATE $\mathcal{T}$ = 0  
    \WHILE{$\mathcal{T} \leq \mathcal{T_{\text{max}}}$}
        \STATE a leaf node $x_i \leftarrow \text{SELECTION}(x_0)$
        \STATE observe state $s$
        \STATE $\Gamma(s) = \text{EXPANSION}(s)$
        \WHILE{$n \leq \mathcal{N}$}
        \STATE $X'$ = SIMULATION($s' \sim \Gamma(s)$)
            \IF{$X'$ is a feasible solution}
            \STATE MIRROR($X'$)
            \ENDIF
            \STATE $\tau(s), r(s)$ = BACKUP($X'$)
        \ENDWHILE
        \STATE y = Eq. (\ref{eq:y})
        \STATE Store $(s, a, y)$ in $\mathcal{R}$
        \STATE Randomly sample a minibatch $\mathcal{B}$ from $\mathcal{R}$
        \STATE Perform a gradient descent step to update $w$
        \STATE $ \mathcal{T} \leftarrow \mathcal{T} + 1$
    \ENDWHILE
    \STATE For every $p$, set $\bar{w} = w$
\ENDFOR
\RETURN $\bar{w}$
\end{algorithmic}
\end{algorithm}

Our algorithm, presented in Algorithm \ref{alg:algorithm}, is designed for training a target network using a given training set $\mathcal{P}$. The input of algorithm is a training set $\mathcal{P}$, and its output is the trained parameter $\bar{w}$ for the target network. Initialization is performed in Steps \ref{alg:line:initM}-\ref{alg:line:root}, where experience buffers and root nodes are set up.
During the iterative process, $T_{\text{max}}$ is limited in steps 4-5, representing iterations for each instance $p$ in the training set $\mathcal{P}$. Step 6 employs the SELECTION process to navigate from the root node $x_0$ to a leaf node $x_i$.
In Steps 7-8, the algorithm observes states $s$ and expands nodes corresponding to $s$, resulting in the set of all child nodes $\Gamma(s)$. Steps 9-15 executed $\mathcal{N}$ iteration. In each iteration, a state $s'$ is randomly selected from $\Gamma(s)$, and a SIMULATION is performed to obtain a variable assignment $X'$. If $X'$ proves to be a feasible solution, a MIRROR operation is executed. The BACKUP operation updates $\tau(s)$ and $r(s)$ for each state in the path from the root to state $s$. From steps 16 to 20, the target $y$ is computed according to Eq. \ref{eq:y}, then stores the experience $(s, a, y)$ in the experience replay buffer $\mathcal{R}$, and updates the online network $Q_w$. Additionally, the iteration counter $\mathcal{T}$ is increased. Finally, for each instance $p$, updated the target network parameters $\bar{w} = w$.


\begin{center}{
\begin{table*}[!ht]
\centering
\fontsize{4}{6}\selectfont
\resizebox{\textwidth}{!}{
\begin{tabular}{llrrrrrrrrr}
\hline
\multicolumn{11}{c}{Random COPs $ \langle m = 2, \gamma = 0.7, \beta = 3, \rho = 0.21 \rangle $ }     \\ \hline
             &          & \multicolumn{3}{c}{n = 20} & \multicolumn{3}{c}{n = 30} & \multicolumn{3}{c}{n = 40} \\ \cline{3-5} \cline{6-8} \cline{9-11}
             & Methods  & Gap  & Time  & Search Node & Gap  & Time  & Search Node & Gap  & Time  & Search Node \\ \hline
             & \textbf{$\Ours$}     & \textbf{0.00\%}         & 1.14s                    & \textbf{1207}                   & \textbf{0.00\%}         & 21.16s           & \textbf{30294}                  & 8.79\%                  & \textbf{3m6s}            & \textbf{284926}                 \\
 & Toulbar2  & \textbf{0.00\%}         & \textbf{1.11s}           & 5891                         & \textbf{0.00\%}         & \textbf{12.39s}          & 124081                   & \textbf{0.00\%}         & 4m5s                     & 1359620                         \\
             & DABP     & 2.94\%                  & 4m27s                    & -                              & 14.21\%                 & 7m19s                    & - & 23.62\%  & 12m32s & - \\
            \hline
\multicolumn{11}{c}{WGCPs $ \langle m = 2, \gamma = 0.7, \beta = 3, \rho = 0 \rangle $}     \\ \hline
             & \textbf{$\Ours$}     & 0.00\%                  & 2.01s                   & \textbf{1892}                   & 0.68\%                  & 37.51s                   & \textbf{53928}                  & 9.14\%                  & \textbf{2m6s}            & \textbf{302918}                 \\
 & Toulbar2 & 0.00\%                  & \textbf{1.11s}           & 10294                           & \textbf{0.00\%}         & \textbf{31.49s}          & 313942                          & \textbf{0.00\%}         & 5m46s                    & 1671902                         \\
             & DABP     & 0.00\%                  & 4.62s                    & -                               & 0.81\%                  & 1m46s                    & -                               & 12.62\%                 & 2m43s                    & -                              \\
                              \hline
\end{tabular}
}
\caption{The performance of different methods for random COPs ($\Delta = 5$) and WGCPs ($\Delta = 1$). }
\label{tab:CopResult}
\end{table*}
}
\end{center}
\section{Experiments}

In this section, we conduct comprehensive experiments on three instance-generated tasks with different parameters to evaluate $\Ours$, especially emphasizing the capability of MCTS to plan variables ordering, thereby reducing the number of nodes required to search for feasible solutions. Specifically, we first introduce the setup of our experiments, followed by the presentation of testing results on different tasks with different distributions, and finally present the analysis of the performance on top-$k$ solutions.

\subsection{Experimental setup}
\textbf{Instance generation.} In our experiments, we consider three standard benchmark types: random Constraint Optimization Problems (COPs), Weighted Graph Coloring Problems (WGCP), and random Constraint Satisfaction Problems (CSPs). To randomly generate tasks, we employ the RB model \cite{xu2007random}, utilizing five parameters $\langle m, n, \alpha, \rho, \beta\rangle$.  It is noteworthy that, for generating CSP instances, an additional parameter $\Delta$ is introduced, and the generation process involves assigning a random cost from $[0, \Delta]$ to each pair of constraints. The specific purposes of these parameters are detailed below:
\begin{itemize}
    \item $m \geq 2$ is the arity of each constraint;
    \item $n \geq 2$ is the number of variables;
    \item $\gamma > 0$ specifies $d$, which is the domain size of each variable, and $d = n^\alpha$;
    \item $\beta > 0$ specifies $e$, which is the number of constraints, and $e = \beta \cdot n \cdot \ln n$;
    \item $\rho \in (0, 1)$ specifies the constraint tightness and $\rho \cdot d^k$ is the number of disallowed tuples for each constraint;
    \item $\Delta$ specifies the difference between the maximum and minimum values in the constraint.
\end{itemize}
Each unique combination of the above parameters delineates a class of CSP instances, which can be regarded as the distribution. The CSP instances employed are all positioned at the phase transition thresholds, where parameter combinations result in the most challenging instances. An advantageous theoretical property of the RB model, distinguishing it from other random CSP models, is its ability to ensure precise phase transitions and instance difficulty at the threshold \cite{xu2007random}. We evaluate our approach across distribution $\langle 2, n, 0.7, 3, 0.21, \Delta \rangle$ for the random CSP and COP tasks, and on a distribution of $\langle 2, n, 0.7, 3, 0, \Delta \rangle$ for the WGCP task. Instances of varying scales are generated by adjusting the values of $n$ and $\Delta$ for different tasks. Specifically, in the case of random COP tasks, we opt for $n = \{20, 30, 40\}$ and $\Delta = 5$, indicating weights taking the values from $\{0,1,2,3,4,5\}$. As for WGCP, we select $n =\{20,30,40\}$ and $\Delta = 1$, normalizing all weights to $[0,1]$. For the random CSP, we choose $n =\{15,20,25,30\}$ and $\Delta = 0$. \\
\textbf{Baselines.} For COPs baselines, we compare our proposed approach with the state-of-the-art COP solvers: \textbf{(1)} DABP with a splitting ratio of 0.95 \cite{deng2022deep} \textbf{(2)} Toulbar2 with a timeout of 1200s \cite{sp2010toulbar2}. For CSPs baselines, we compare with the state-of-the-art CSP solvers such as DRL \cite{DBLP:journals/eaai/SongC0X022}. Additionally, we evaluate against four classic hand-crafted variable ordering heuristics commonly employed in various CSP solvers: MinDom \cite{haralick1980increasing}, Impact \cite{refalo2004impact}, Dom/Ddeg \cite{bessiere1996mac} and Dom/Tdeg \cite{li2016improving}, which are utilized as features in our methods. \\
\textbf{Implementation details.} For our GNN model, we fix the embedding dimension at $p = 128$, and all MLPs have $L=3$ layers with hidden dimension $64$ and rectified linear units (RELU) as activation functions. The embeddings are updated for $K = 5$ iterations. Our model is implemented with the PyTorch Geometric framework \cite{fey2019fast} and trained with the Adam optimizer \cite{kinga2015method} using the learning rate of $5 \times 10^{-5}$ and mini-batch size $\mathcal{B} = 128$. The size of experience replay is $\mathcal{M} = 1$M. The frequency of updating the target network is $e_u = 10$. 
We employ $\mathcal{P} = 100$ instances, and the simulation $\mathcal{T}_\text{max}$ is limited to $10000$ with $\mathcal{N} = 10$, indicating that each expanded node will be simulated $10$ times. For testing, we impose a cutoff limit of $5 \times 10^5$ search nodes for our policies. The implementation of our approach is carried out in Python. All experiments are conducted on an Intel(R)-E5-2637 with RTX 1080 GPU (11GB memory) and 251GB memory.
\begin{center}{
\begin{table}[!ht]
\scriptsize
\resizebox{\linewidth}{!}{
\begin{tabular}{llrrl}
                        \hline
                        \multicolumn{5}{c}{Random CSPs $\langle m = 2, \gamma = 0.7, \beta = 3, \rho = 0.21, \Delta = 0 \rangle$}\\
                        \hline
                        & Methods    & \multicolumn{2}{l}{\# Search Nodes}                         & \#cutoff \\ \cline{3-4} 
                        &              & Average                     & Reduction                     &          \\
                        \hline
                        & \textbf{$\Ours$}         & \textbf{21.97}              & -                 & -        \\
                        & DRL          & \textbf{22.21}              & -                             & -        \\
n = 15                  & Dom/Tdeg     & \textbf{22.81}              & -                             & -        \\
                        & Dom/Ddeg     & 23.05                       & 4.67\%                        & -        \\
                        & MinDom       & 33.57                       & 34.55\%                       & -        \\
                        & Impact       & 272.51                      & 91.93\%                       & -        \\
                        \hline
                        & \textbf{$\Ours$}         & \textbf{273.86}             & -                 & -        \\
                        & DRL          & 291.30                      & 5.99\%                        & -        \\
n = 25                  & Dom/Tdeg     & 320.19                      & 14.25\%                       & -        \\
                        & Dom/Ddeg     & 347.78                      & 21.25\%                       & -        \\
                        & MinDom       & 799.54                      & 65.75\%                       & -        \\
                        & Impact       & 69885.05                    & 99.61\%                       & -        \\
                        \hline
                        & \textbf{$\Ours$  (n = 25)} & \textbf{1149.52}            & -               & -        \\
                        & DRL  (n = 25)  & 1237.93                     & 7.14\%                      & -        \\
n = 30                  & Dom/Tdeg       & 1350.92                     & 14.91\%                     & -        \\
                        & Dom/Ddeg     & 1523.92                     & 24.57\%                       & -        \\
                        & MinDom       & 4160.78                     & 72.37\%                       & -        \\
                        & Impact       & 318862.66                   & 99.63\%                       & 25      \\
                        \hline
                        & \textbf{$\Ours$ (n = 25)} & \textbf{21547.30}           & -                & -        \\
                        & DRL (n = 25)  & 23684.04                    & 9.02\%                       & -        \\
n = 40                  & Dom/Tdeg     & 26861.89                    & 19.78\%                       & -        \\
                        & Dom/Ddeg     & 31807.81                    & 32.26\%                       & -        \\
                        & MinDom       & 136405.52                   & 84.20\%                       & 2       \\
                        & Impact       & 491872.80                   & 95.62\%                       & 49     \\
                                            \hline
\end{tabular}
}
\caption{Test CSPs result on the same distribution used in training. The cutoff indicates how many instances have more search nodes than the $5 \times 10^5$.}
\label{tab:CSPresult}
\end{table}
}
\end{center}
\subsection{Experiment Results}
In this section, we discuss the performance of the $\Ours$ during training. Specifically, we train the agent on the instances with different distributions. We use three measures to assess our proposed method, including (1) the number of search nodes, which represents the search nodes required to find the feasible solutions we aim to minimize, (2) the gap indicates how close a solution is to the optimal solution, and (3) the execution time, indicating the time taken to find a feasible solution.\\
\textbf{Performance Comparison for random COPs.} Table \ref{tab:CopResult} shows the performance of three different methods when applied to Random COPs and WGCPs.
We make the following observations: \textbf{(1)} In the case of random COP, especially for $n=20$, $\Ours$ performs exceptionally well in terms of both GAP and Time, with a GAP of 0.00$\%$, indicating that we have found the optimal solution. When compared to Toulbar2, although our time is slightly longer by $0.03$ seconds, our search node number is approximately $\textbf{4.9}$ times less than Toulbar2, suggesting that we can find the solution with fewer search nodes. Furthermore, as $n$ increases to $40$, our method is not only faster in terms of time compared to Toulbar2 but also has significantly fewer search nodes than Toulbar2, which demonstrates the superiority of our method. In both gap and time measures, our approach outperforms DABP. It is worth noting that DABP is an end-to-end approach therefore there are no search nodes, \textbf{(2)} On the other hand, Toulbar2 achieves optimal solutions for all cases but explores a greater number of nodes in each case compared to our $\Ours$ method. This is expected as Toulbar2 systematically searches the entire solution space to obtain the optimal solution,
and \textbf{(3)} for the WGCPs task, $\Ours$ performance follows a similar trend to the effect on the COPs task. Again our search nodes are still much lower than the Toulbar2 method.
The conclusion drawn from this table is that our model performs better than both DABP and Toulbar in complex and large-scale problems. It is also evident that our model can find a near-optimal solution using a relatively small number of search nodes, which aligns with our motivation. \\
\textbf{Performance Comparison for random CSPs.} Utilizing the zero objective values in the original constraint features, we established a CSP problem-solving model. Table \ref{tab:CSPresult} shows the performance of our algorithm in tackling random CSP problems of varying scales. Overall, $\Ours$ exhibits the fewest search nodes across all scales. Additionally, for scales $n = 30$ and $n = 40$, $\Ours$ and DRL employed models trained with $n = 25$, denoted as $\Ours$$(n = 25)$ and DRL$(n = 25)$, respectively. It is noteworthy that, even when using $\Ours$$(n=25)$ at $n = 30$ and $n = 40$, the number of search nodes consistently remains lower than other classically designed models. Due to the independence of our initial feature density on the problem scale, our model demonstrates a certain degree of universality, indicating a level of generalization. This emphasizes the superiority of our method in reducing the number of search nodes, especially performing well in small-scale problems. 


\begin{figure}[!ht]
    \begin{minipage}{0.5\textwidth}
    \centering
    \includegraphics[width=1.0\linewidth]{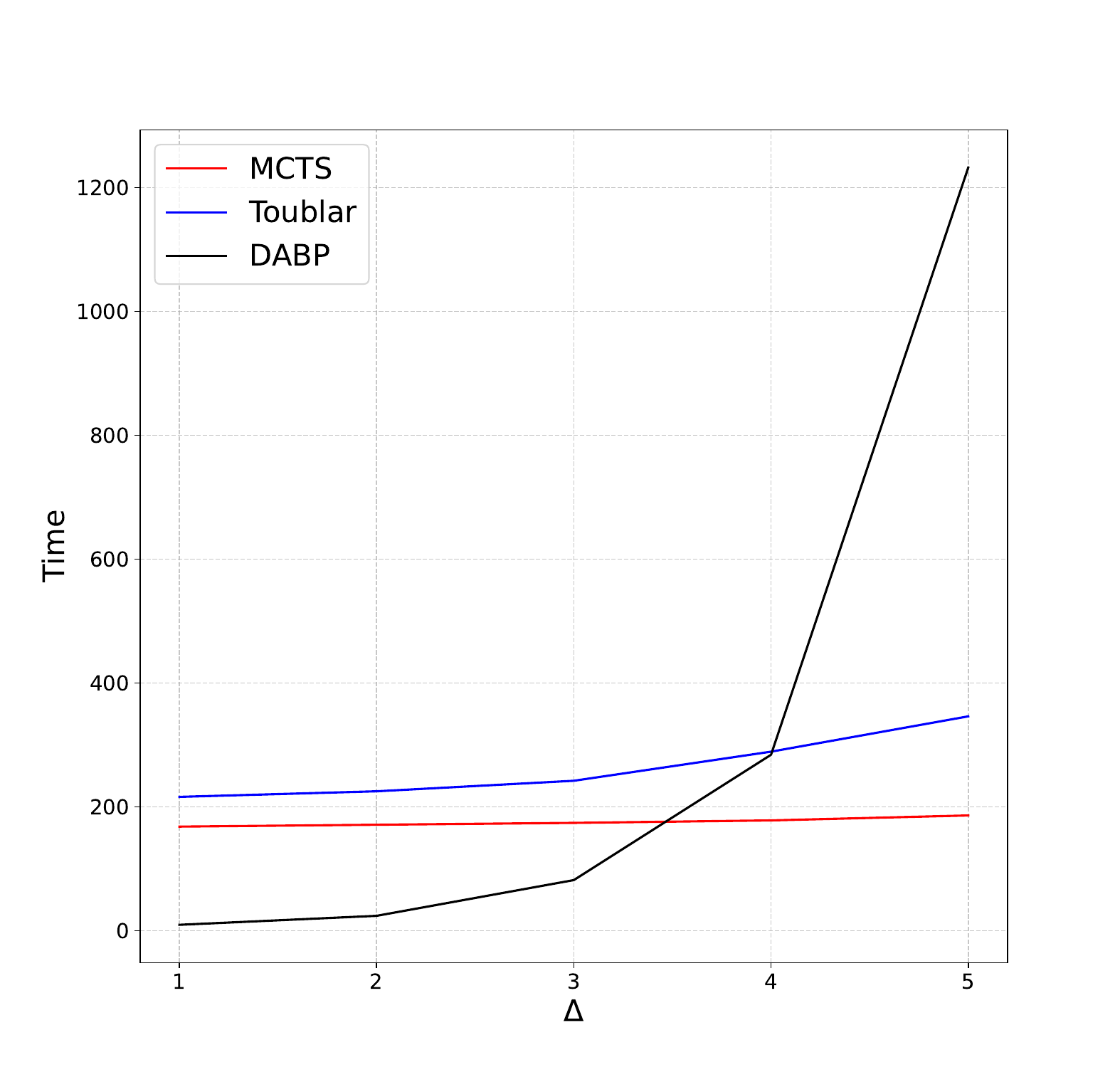} 
    \caption{Solution times of the three methods at distribution $\langle m = 2, n = 40, \gamma = 0.7, \beta = 3, \rho = 0.21, \Delta = 5 \rangle$.}
    \label{fig:figure4}
    \end{minipage}
\end{figure}

\subsection{Additional analyses.}
\textbf{Solution time at diffrent $\Delta$}
Figure \ref{fig:figure4} shows the execution time of the three methods for different $\Delta$ at the same distribution. Focusing only on the scenario of finding a high-quality solution, our $\Ours$ running time is shorter. Moreover, the solution time of our $\Ours$ method does not change significantly as the delta increases.  In particular, the solution time of $\Ours$ is shortest when the delta is large.\\
\textbf{Average gap of top-$k$ solutions.} 
In Table \ref{tab:topk-gap}, we show the optimal gap that can be achieved for the top-$k$ solutions for the random COPs task, showing more clearly the efficiency of our model in finding high-quality feasible solutions that may be close to the optimal or best solution.
From Table \ref{tab:topk-gap}, it is evident that our method achieves an average gap of less than 30\% in the first obtained solution, as previously indicated in Table \ref{tab:CopResult}, showcasing the significantly reduced number of nodes required compared to the Toulbar2 solver. This performance is noteworthy, with an average gap of less than 17.63\% in the top 5 solutions and the ability to find optimal solutions in simple instances within the top 20 solutions.

\begin{table}[!ht]
\centering
\scriptsize
\resizebox{\linewidth}{!}{
\begin{tabular}{llrrrr}
\hline
\multicolumn{6}{l}{Random COPs $\langle m = 2, \gamma = 0.7, \beta = 3, \rho = 0.21, \Delta = 5 \rangle$} \\
\hline
       & n = 20                     & n = 25 & n = 30 & n = 35 & n = 40 \\
\hline
k = 1  & 5.71\%                     & 10.53\%                    & 13.81.\%                   & 19.44\%                    & 28.92\%                    \\
k = 5  & 2.28\%                     & 3.59\%                     & 6.72\%                     & 13.86\%                    & 17.63\%                    \\
k = 10 & 1.42\%                     & 1.82\%                     & 3.48\%                     & 8.65\%                     & 15.24\%                    \\
k = 20 & 0.00\% & 0.00\%                     & 0.00\%                     & 3.24\%                     & 9.59\%                      \\
\hline
\end{tabular}
}
\caption{Average gap of top-k solutions over different $n$.}
\label{tab:topk-gap}
\end{table}

\section{Final Remarks} 
In this study, we introduce an MCTS framework called $\Ours$, designed for training a neural network as a heuristic function. The aim is to optimize the recommendation order of variables when solving COP. Experiments Random COP demonstrate that our algorithm consistently identifies high-quality solutions within the top 5 feasible solutions. Through experimentation with Random CSP, we validate that our algorithm requires minimal node exploration to find solutions. However, an important limitation is the significant computational cost incurred during the inference stage. In spite of the efficiency of $\Ours$ in terms of node search, the overall computational time has not decreased significantly. It is also possible to further optimize the neural network by reducing its size to improve the speed of inference in the future.
\newpage
\bibliography{aaai24}

\begin{thebibliography}{26}
\providecommand{\natexlab}[1]{#1}

\bibitem[{Abe et~al.(2019)Abe, Xu, Sato, and Sugiyama}]{abe2019solving}
Abe, K.; Xu, Z.; Sato, I.; and Sugiyama, M. 2019.
\newblock Solving np-hard problems on graphs with extended alphago zero.
\newblock \emph{arXiv preprint arXiv:1905.11623}.

\bibitem[{Achterberg and Wunderling(2013)}]{achterberg2013mixed}
Achterberg, T.; and Wunderling, R. 2013.
\newblock Mixed integer programming: Analyzing 12 years of progress.
\newblock In \emph{Facets of combinatorial optimization: Festschrift for martin
  gr{\"o}tschel}, 449--481. Springer.

\bibitem[{Bertsekas(2014)}]{bertsekas2014constrained}
Bertsekas, D.~P. 2014.
\newblock \emph{Constrained optimization and Lagrange multiplier methods}.
\newblock Academic press.

\bibitem[{Bessiere and R{\'e}gin(1996)}]{bessiere1996mac}
Bessiere, C.; and R{\'e}gin, J.-C. 1996.
\newblock MAC and combined heuristics: Two reasons to forsake FC (and CBJ?) on
  hard problems.
\newblock In \emph{International Conference on Principles and Practice of
  Constraint Programming}, 61--75. Springer.

\bibitem[{Deng et~al.(2022)Deng, Kong, Liu, and An}]{deng2022deep}
Deng, Y.; Kong, S.; Liu, C.; and An, B. 2022.
\newblock Deep Attentive Belief Propagation: Integrating Reasoning and Learning
  for Solving Constraint Optimization Problems.
\newblock \emph{Advances in Neural Information Processing Systems}, 35:
  25436--25449.

\bibitem[{Farinelli et~al.(2008)Farinelli, Rogers, Petcu, and
  Jennings}]{farinelli2008decentralised}
Farinelli, A.; Rogers, A.; Petcu, A.; and Jennings, N.~R. 2008.
\newblock Decentralised coordination of low-power embedded devices using the
  max-sum algorithm.

\bibitem[{Fey and Lenssen(2019)}]{fey2019fast}
Fey, M.; and Lenssen, J.~E. 2019.
\newblock Fast graph representation learning with PyTorch Geometric.
\newblock \emph{arXiv preprint arXiv:1903.02428}.

\bibitem[{Gleixner et~al.(2021)Gleixner, Hendel, Gamrath, Achterberg, Bastubbe,
  Berthold, Christophel, Jarck, Koch, Linderoth et~al.}]{gleixner2021miplib}
Gleixner, A.; Hendel, G.; Gamrath, G.; Achterberg, T.; Bastubbe, M.; Berthold,
  T.; Christophel, P.; Jarck, K.; Koch, T.; Linderoth, J.; et~al. 2021.
\newblock MIPLIB 2017: data-driven compilation of the 6th mixed-integer
  programming library.
\newblock \emph{Mathematical Programming Computation}, 13(3): 443--490.

\bibitem[{Haralick and Elliott(1980)}]{haralick1980increasing}
Haralick, R.~M.; and Elliott, G.~L. 1980.
\newblock Increasing tree search efficiency for constraint satisfaction
  problems.
\newblock \emph{Artificial intelligence}, 14(3): 263--313.

\bibitem[{Kinga, Adam et~al.(2015)}]{kinga2015method}
Kinga, D.; Adam, J.~B.; et~al. 2015.
\newblock A method for stochastic optimization.
\newblock In \emph{International conference on learning representations
  (ICLR)}, volume~5, 6. San Diego, California;.

\bibitem[{Kschischang, Frey, and Loeliger(2001)}]{kschischang2001factor}
Kschischang, F.~R.; Frey, B.~J.; and Loeliger, H.-A. 2001.
\newblock Factor graphs and the sum-product algorithm.
\newblock \emph{IEEE Transactions on information theory}, 47(2): 498--519.

\bibitem[{Li et~al.(2016)Li, Liang, Zhang, Guo, Xu, and Li}]{li2016improving}
Li, H.; Liang, Y.; Zhang, N.; Guo, J.; Xu, D.; and Li, Z. 2016.
\newblock Improving degree-based variable ordering heuristics for solving
  constraint satisfaction problems.
\newblock \emph{Journal of Heuristics}, 22: 125--145.

\bibitem[{Lodi and Zarpellon(2017)}]{lodi2017learning}
Lodi, A.; and Zarpellon, G. 2017.
\newblock On learning and branching: a survey.
\newblock \emph{Top}, 25: 207--236.

\bibitem[{Loth et~al.(2013)Loth, Sebag, Hamadi, and
  Schoenauer}]{loth2013bandit}
Loth, M.; Sebag, M.; Hamadi, Y.; and Schoenauer, M. 2013.
\newblock Bandit-based search for constraint programming.
\newblock In \emph{Principles and Practice of Constraint Programming: 19th
  International Conference, CP 2013, Uppsala, Sweden, September 16-20, 2013.
  Proceedings 19}, 464--480. Springer.

\bibitem[{Modi et~al.(2005)Modi, Shen, Tambe, and Yokoo}]{modi2005adopt}
Modi, P.~J.; Shen, W.-M.; Tambe, M.; and Yokoo, M. 2005.
\newblock ADOPT: Asynchronous distributed constraint optimization with quality
  guarantees.
\newblock \emph{Artificial Intelligence}, 161(1-2): 149--180.

\bibitem[{Refalo(2004)}]{refalo2004impact}
Refalo, P. 2004.
\newblock Impact-based search strategies for constraint programming.
\newblock In \emph{Principles and Practice of Constraint Programming--CP 2004:
  10th International Conference, CP 2004, Toronto, Canada, September 27-October
  1, 2004. Proceedings 10}, 557--571. Springer.

\bibitem[{Rollon and Larrosa(2014)}]{rollon2014decomposing}
Rollon, E.; and Larrosa, J. 2014.
\newblock Decomposing utility functions in bounded max-sum for distributed
  constraint optimization.
\newblock In \emph{International conference on principles and practice of
  constraint programming}, 646--654. Springer.

\bibitem[{Sabharwal, Samulowitz, and Reddy(2012)}]{sabharwal2012guiding}
Sabharwal, A.; Samulowitz, H.; and Reddy, C. 2012.
\newblock Guiding combinatorial optimization with UCT.
\newblock In \emph{Integration of AI and OR Techniques in Contraint Programming
  for Combinatorial Optimzation Problems: 9th International Conference, CPAIOR
  2012, Nantes, France, May 28--June1, 2012. Proceedings 9}, 356--361.
  Springer.

\bibitem[{Satomi et~al.(2011)Satomi, Joe, Iwasaki, and Yokoo}]{satomi2011real}
Satomi, B.; Joe, Y.; Iwasaki, A.; and Yokoo, M. 2011.
\newblock Real-time solving of quantified csps based on monte-carlo game tree
  search.
\newblock In \emph{Twenty-Second International Joint Conference on Artificial
  Intelligence}. Citeseer.

\bibitem[{Scavuzzo et~al.(2022)Scavuzzo, Chen, Ch{\'e}telat, Gasse, Lodi,
  Yorke-Smith, and Aardal}]{scavuzzo2022learning}
Scavuzzo, L.; Chen, F.; Ch{\'e}telat, D.; Gasse, M.; Lodi, A.; Yorke-Smith, N.;
  and Aardal, K. 2022.
\newblock Learning to branch with tree mdps.
\newblock \emph{Advances in Neural Information Processing Systems}, 35:
  18514--18526.

\bibitem[{Silver et~al.(2017)Silver, Schrittwieser, Simonyan, Antonoglou,
  Huang, Guez, Hubert, Baker, Lai, Bolton et~al.}]{silver2017mastering}
Silver, D.; Schrittwieser, J.; Simonyan, K.; Antonoglou, I.; Huang, A.; Guez,
  A.; Hubert, T.; Baker, L.; Lai, M.; Bolton, A.; et~al. 2017.
\newblock Mastering the game of go without human knowledge.
\newblock \emph{nature}, 550(7676): 354--359.

\bibitem[{Song et~al.(2022)Song, Cao, Zhang, Xu, and
  Lim}]{DBLP:journals/eaai/SongC0X022}
Song, W.; Cao, Z.; Zhang, J.; Xu, C.; and Lim, A. 2022.
\newblock Learning variable ordering heuristics for solving Constraint
  Satisfaction Problems.
\newblock \emph{Eng. Appl. Artif. Intell.}, 109: 104603.

\bibitem[{SP, SP, and SP(2010)}]{sp2010toulbar2}
SP, M.~S.; SP, F.~H.; and SP, E.~R. 2010.
\newblock ToulBar2, an open source exact cost function network solver.

\bibitem[{Xu et~al.(2007)Xu, Boussemart, Hemery, and Lecoutre}]{xu2007random}
Xu, K.; Boussemart, F.; Hemery, F.; and Lecoutre, C. 2007.
\newblock Random constraint satisfaction: Easy generation of hard (satisfiable)
  instances.
\newblock \emph{Artificial intelligence}, 171(8-9): 514--534.

\bibitem[{Xu et~al.(2019)Xu, Hu, Leskovec, and
  Jegelka}]{DBLP:conf/iclr/XuHLJ19}
Xu, K.; Hu, W.; Leskovec, J.; and Jegelka, S. 2019.
\newblock How Powerful are Graph Neural Networks?
\newblock In \emph{7th International Conference on Learning Representations,
  {ICLR} 2019, New Orleans, LA, USA, May 6-9, 2019}. OpenReview.net.

\bibitem[{Xu, Kadam, and Lieberherr(2021)}]{xu2021first}
Xu, R.; Kadam, P.; and Lieberherr, K. 2021.
\newblock First-order problem solving through neural mcts based reinforcement
  learning.
\newblock \emph{arXiv preprint arXiv:2101.04167}.

\end{thebibliography}
\end{document}